# Semantic Prosody in Machine Translation: the English-Chinese Case of Passive Structures

**Xinyue Ma**[1,2], **Pol Pastells**[1,2], **Mireia Farrús**[1,2], **Mariona Taulé**[1,2]
[1]Centre de Llenguatge i Computacio (CLiC), Universitat de Barcelona, Spain
[2]Institut de Recerca en Sistemes Complesos (UBICS), Universitat de Barcelona, Spain
`{maxinyue, pol.pastells, mfarrus, mtaule}@ub.edu`

## Abstract

Semantic prosody is a collocational meaning formed through the co-occurrence of a linguistic unit and a consistent series of collocates, which should be treated separately from semantic meaning. Since words that are literal translations of each other may have different semantic prosody, more attention should be paid to this linguistic property to generate accurate translations. However, current machine translation models cannot handle this problem. To bridge the gap, we propose an approach to teach machine translation models about semantic prosody of a specific structure. We focus on Chinese BEI passives and create a dataset of English-Chinese sentence pairs with the purpose of demonstrating the negative semantic prosody of BEI passives. Then we fine-tune OPUS-MT, NLLB-600M and mBART50 models with our dataset for the English-Chinese translation task. Our results show that fine-tuned MT models perform better on using BEI passives for translating unfavourable content and avoid using it for neutral and favourable content. Also, in NLLB-600M, which is a multilingual model, this knowledge of semantic prosody can be transferred from English-Chinese translation to other language pairs, such as Spanish-Chinese.

## 1 Introduction

The notion of semantic prosody was first proposed in corpus linguistics in the 1990s. It was Sinclair (1987) who first noticed the phenomenon that some words or phrases tend to co-occur with unpleasant events, such as *HAPPEN* and *SET IN*. This property was later named Semantic Prosody (an analogy to Firth (1964)'s concept of phonological prosody) by Louw (1993) and defined as "a consistent aura of meaning with which a form is imbued by its collocates". Another well-known example is *CAUSE* as a verb (Stubbs, 1995). As shown in Figure 1, entities "caused" are generally considered undesirable: rupture, death, pain, misery, damage, and diseases.

Sinclair (1996) considers semantic prosody to be on the "pragmatic side of the semantics/pragmatics continuum" and can reveal the speaker or writer's attitude. However, semantic prosody cannot be precisely perceived by individual native speakers through intuition (Louw, 1993; McEnery et al., 2006). A statistical analysis of corpus data is needed to correctly interpret the semantic prosody of a linguistic unit (a node), since corpus data can provide multiple usage patterns across many speakers and reflect the general intuition of native speakers (Stubbs, 2001; McEnery et al., 2006; Stewart, 2010). The co-occurrence of a node and a predominantly positive/negative context can foster subtle associative meaning, and the node is then considered to have positive/negative semantic prosody.

A word in English probably has a synonym in Chinese with the same semantic meaning. However, the two words do not necessarily have the same semantic prosody due to language variation. For example, the semantic meaning of the English word *BECAUSE* and the Chinese word "由于 *YOUYU*" are equivalent, but their semantic prosodies are not. *BECAUSE* has a neutral semantic prosody, while *YOUYU* has a negative one (Wu and Lan, 2019). Another example is *INSIST ON*, which is often used to describe annoying stubbornness, while its literal translation in Chinese "坚持 *JIANCHI*" has positive semantic prosody. This divergence in semantic prosody has even caused English learners who are native Mandarin Chinese speakers to use *INSIST ON* to mean encouragement. At the same time, this usage has never appeared in COCA[1] (Dong, 2020).

During the translation process, semantic prosody should be considered in order to achieve semantic/pragmatic equivalence and avoid missing information. Since it is difficult for individual non-native speakers of a language to correctly perceive semantic prosody with intuition, even professional human translators may fail to convey it into the target text, as they usually translate from their second language (L2) into their mother tongue. In both translation pedagogy and L2 learning, semantic prosody awareness has been considered to be important and of value to learners (Stewart, 2009; McGee, 2012).

---
[1]Available at: `https://www.english-corpora.org/coca/`

| 1 | he trial judge noted that the operative intervention | caused | the **rupture** -- not the failure to earlier diagnose |
| 2 | inutes each and where even a slightest error would | cause | **death**. Because that would NOT be fun. Also, if the |
| 3 | omfort. They are specially designed to relieve **pain** | caused | by bruxing or clenching. # If you believe you grin |
| 4 | enters of power. They have spilled more blood and | caused | more **misery** than any other force. They have canoni |
| 5 | and Limn. # The drop in rainfall and soil saturation | caused | **landslides** in the canton of Turrialba, Paraiso, Ji |
| 6 | system cope with the impending public health crisis | caused | **Alzheimer's disease** and related dementia will be p |
| 7 | , recently elected Socialist Party leader Harlem Dsir | caused | **a stir** when he reminded the crowd of President Fra |
| 8 | SE by Vivian Sharpe. It explains the **damage** that is | caused | by sexual abuse. It reminds us why it is so import |
| 9 | d how autism is the the result of **" brain damage "** | caused | by vaccines; then the confabulation begins that re |
| 10 | search engines have her name blocked. **Diabetes** is | caused | by parasites? So, if she didn't have an accumulati |
| 11 | of Somalia have made the waters their home and | have caused | **extensive damage** as a result. Shipping industries |
| 12 | be videos. # And with **all the astrosities** this animal | has caused | , if he does not get impeached, tried. sentenced, |

Figure 1: Concordance lines of *CAUSE* in the Corpus of Contemporary English in the United States (COCA).

Inequivalence of semantic prosody is also a problem for Machine Translation (MT), yet little attention has been paid to it. If we ask a model to translate "I was praised by my teacher" into Chinese, Google Translate, ChatGPT 4 and DeepL would give the following translations:

**Google Translate & ChatGPT 4:**

(1) 我 被 老师 表扬 了。
1SG BEI teacher praise PERF
'I was praised by my teacher.'

**DeepL:**

(2) 我 受到 了 老师的 表扬。
1SG SHOUDAO[2] PERF teacher's praise-VN
'I was praised by my teacher.'

These are two different literal translations in passive voice. In English, passive structures such as "be + past participle" (hereafter referred to as "*BE passive*") mainly occur in neutral contexts (Xiao et al., 2006). However, these structures have negative semantic prosody in Mandarin Chinese (Wu, 2022; Dong et al., 2023). As the standard and most common passive structure in Chinese, the "被 BEI + verb" structure (hereafter referred to as *BEI passive*) has obvious negative prosody, which is not an adequate option for translating a sentence talking about "being praised" or any other favourable situation.

---

[2]SHOUDAO ("be given or undergo") is a delexicalized verb (light verb) and marks passive voice (Cai et al., 2019). In this light verb pattern, the main semantic content of the predicate is provided not by it, but by its action nominal complement，verbal noun 表扬 BIAOYANG ("praising"). The subject is the patient here.

To the best of our knowledge, no previous studies have attempted to teach an MT model about semantic prosody awareness. In this paper, we propose a method to incorporate semantic prosody information regarding a specific structure, namely BEI passive of Mandarin Chinese, into a Sequence-to-Sequence machine translation model. To this end, we introduce a dataset created to demonstrate the negative semantic prosody of BEI passives, which is later used to fine-tune MT models.

The main contributions of this paper are as follows:

1. We propose a method to teach semantic prosody awareness of a specific structure to Seq2Seq MT models, that is, fine-tuning them with a dataset that explicitly demonstrates the semantic prosody of the node.

2. We create a first-of-its-kind English-Chinese parallel dataset on semantic prosody. All sentence pairs are manually selected to illustrate the fact that Chinese BEI passive has negative semantic prosody and is the appropriate translation of English BE passives only if the context is unfavourable.

3. We employ a probing task to validate the idea that fine-tuned models contain more information that helps to decide whether a passive structure should be used in translation.

4. We achieve better performance with our fine-tuned models on translating passives while maintaining original BLEU, chrF2 and CometKiwi scores on Flores+ and Tatoeba datasets.

## 2 Related Work

Research on semantic prosody has developed in multiple directions, ranging from theoretical discussions to empirical studies in contrastive linguistics and translation. In this section, we first outline the key concepts related to semantic prosody, and then examine how it has been applied in cross-linguistic studies and translation research.

### 2.1 Semantic Prosody

Semantic prosody is a collocational meaning formed through the co-occurrence of a node and a consistent series of collocates (Louw, 2000). Another related concept is semantic preference. A node can display its semantic preference by co-occurring with several items from a specific semantic set, which can contain favourable and unfavourable items at the same time. According to Partington (2004) semantic preference and semantic prosody have different operating scopes: the former relates the node item to another item from a particular semantic set whereas the latter can affect wider stretches of text. Semantic preference is a feature of the collocates, while semantic prosody is a feature of the node word. For example, *ABSOLUTELY* shows a semantic preference for words with a strong or superlative sense, such as *DELIGHTED, ENCHANTING, SPLENDID, PREPOSTEROUS, APPALLING, INTOLERABLE* (Partington, 1991). Both semantic prosody and semantic preference are established through collocates, and semantic preference "contributes powerfully" to building semantic prosody (Partington, 2004).

The definition of semantic prosody has not been undisputed over the last thirty years. Sardinha (2000) and Stubbs (2001) seem to consider that semantic prosody and connotation are synonymous, while Louw (2000) takes a different view. According to the *Collins Cobuild English Dictionary* definition of connotation, "the connotations of a particular word or name are the ideas or qualities which it makes you think of" (Sinclair, 1995). For example, *URCHIN* has a connotation of mischievousness. Louw (2000) takes connotation as "a form of schematic knowledge of repeatable events, e.g., what urchins do, where they live, their financial means or lack of it and how they behave, etc.". At the same time, semantic prosody is more contingent to collocates and requires corpus data to pin it down. It should be noted that although semantic prosody and connotation share the property of being attitudinal, the concealed quality is more fundamental to semantic prosody than it is to connotation (Stewart, 2010), which is far more accessible and can be learned in daily life.

### 2.2 Semantic Prosody in Contrastive Linguistics and Translation

In the field of contrastive linguistics, many case studies have been conducted on various language pairs of Indo-European languages and Sino-Tibetan languages. Xiao and McEnery (2006) compared the prosodies of near-synonyms across English and Chinese, and Sardinha (2000) analyzed English and Portuguese. Both studies conclude that the collocational behaviour and semantic prosodies of near-synonyms are unpredictable across the two language pairs, sometimes appearing similar and other times distinct. As for recent works, Wu and Lan's study on *BECAUSE* and 由于 *YOUYU* (Wu and Lan, 2019), and Dong's study about *INSIST ON* and 坚持 *JIANCHI* (Dong, 2020) also validate this observation. Partington (1998) claims that perfect equivalents across English and Italian are rare between because even words and expressions that are "look-alikes" (e.g., English *CORRECT* vs. Italian *CORRETTO*) may have very different lexical environments. Furthermore, there are many case studies discussing the more appropriate translation of a certain word or phrase (see example (3)). Wang and Ge (2021) claim that considering the negative semantic prosody of "It is what it is", (3-b) is a better translation than (3-a), since the former also has negative semantic prosody whereas the latter mainly appears in a neutral context.

(3) **Source text:** It is what it is.
   a. 情况就是这样。
      'This is the situation.'
   b. 事已至此。
      'The matter has come to this.'

Currently, although the application of semantic prosody to translation, translation pedagogy and L2 learning is the subject of research, yet no study has tried to appreciate it for improving MT performance. Considering its importance in translation equivalence, teaching models about semantic prosody awareness is a feasible way to improve MT translation performance.

## 3 Dataset Description

We propose teaching machine translation models about the negative semantic prosody of Chinese BEI passives by fine-tuning them with a dataset focusing on this structure, (hereafter referred to as the BEI dataset). Detailed information on the dataset is presented in this section.

### 3.1 Linguistic Structures

#### 3.1.1 BEI Passives

The BEI dataset was created to explicitly demonstrate the negative semantic prosody of a passive structure in Chinese (*BEI* + verb) to a translation model. *BEI* is a grammatical passive marker without a concrete semantic meaning and is the most frequently used one among all passive markers. In Chinese, an active sentence can be turned into a passive one through adding a passive marker, and switching the subject and object, that is, the patient becomes the subject. Example (4) shows how such change is made.

(4) a. 张三　打　了　李四
Zhangsan beat PERF Lisi
'Zhangsan beat Lisi'

b. 李四**被** 张三　打　了
Lisi **BEI** Zhangsan beat PERF
'Lisi was beaten by Zhangsan'

The frequency, genre distribution, and semantic prosody of the passive voice differ in Chinese and English. Through an analysis of a corpus of recent materials (literature, news, and papers from January 1st to October 20th, 2021), Dong et al. (2023) reveals that the passive voice is approximately eight times more common in English than in Chinese. In English, it primarily conveys *neutral* content and is more frequent in news and academic articles, which require objectivity, than in novels. In contrast, Chinese uses the passive voice mainly for *negative* content, with little variation across genres, as illustrated in example (5).

(5) 家珍**被**拖出去时，双手紧紧捂着凸起的肚子，那里面有我的儿子呵。

'As Jiazhen **was carried out**, her hands firmly clasped her protruding belly, which held my son.'

The frequency of BEI passives in fiction is higher than in other genres, appearing 153 times per 100,000 words in literary texts, while only 94 times 100,000 in news texts and even less frequently in scientific papers and miscellaneous texts. Meanwhile, the semantic prosody of BEI passives is also the most negative in literary texts, compared to that in other genres. In 66% of cases, the BEI passive has negative collocates, whereas the percentage in news text is 51.5% (Xiao et al., 2006). The frequency of passive sentences in Chinese translated fiction is lower and the semantic prosody of BEI passives is more negative than in Chinese original fiction, showing a tendency toward domestication in translation (Jia, 2010). Considering these facts, the dataset was created only with literary texts.

It should also be noted that although the passive voice is mainly used for unfavourable events in Chinese, its usage in a positive context is not entirely nonexistent. As with the active voice, it can be used in any context, positive and negative alike.

#### 3.1.2 BE Passives

The structure *BE* + past participle can be considered the norm for English passives (Xiao et al., 2006) and is the most frequent passive structure used in English. BE passives appeared 9,908 times in FLOB (Freiburg-LOB corpus[3], an update of the Lancaster-Oslo-Bergen corpus of British English that contains texts published between 1991 and 1992), while GET passives appeared only 59 times. Thus, the BE passive is the structure we looked for when collecting sentence pairs.

English BE passives and Chinese BEI passives show great divergence in semantic prosody. According to Xiao and McEnery (2006), unlike BEI passives, 80% of BE passives in FLOB and BNCdemo (a demographic sampled component of the British National Corpus[4], the World edition) express neutral content.

### 3.2 Dataset Creation

Our dataset contains 900 English-Chinese parallel sentences manually selected from the fiction genre of The Babel English-Chinese Parallel Corpus (244,696 words in total) created by Richard Xiao, and from the China English-Chinese Parallel Corpus-Core (CECPC-Core, 5,499,591 words in total) created by Kefei Wang of BFSU [5].

---
[3] Available at: https://clarino.uib.no/korpuskel/corpora
[4] Available at: http://www.natcorp.ox.ac.uk/
[5] All available at the CQPweb of Beijing Foreign Studies University: http://114.251.154.212/cqp/.

### 3.3 Dataset Analysis

In the BEI dataset, the source texts contain BE passives in all persons and tenses. There are two subsets created with different requirements, namely positive evidence and negative evidence of the usage of BEI passives in the translation of English BE passives:

**Positive evidence:** 476 sentence pairs in which English BE passives are translated to Chinese BEI passives by a human translator and express negative content. They are selected to reinforce the relation between BEI passives and negativity.

**Negative evidence:** 424 English BE passives translated into Chinese with active voice by human translators, and the corresponding Chinese translation. This subset is intended to attenuate the degree of correspondence between the two passives, so that models may use more active voice when translating BE passives into Chinese.

## 4 Experimental Setup and Analysis

In this section, we describe our experiments teaching MT models about the negative semantic prosody of BEI passives. We also present the experimental setup for fine-tuning and probing, a comprehensive analysis of the evaluation and our results.

We experimented with three Sequence to Sequence (Seq2Seq) MT models: *Helsinki-NLP/opus-mt-en-zh* (Tiedemann and Thottingal, 2020; Tiedemann et al., 2023), *facebook/nllb-200-distilled-600M* (Costa-Jussà et al., 2022) and *facebook/mbart-large-50-many-to-many-mmt* (Tang et al., 2020)—hereinafter referred to as OPUS-MT, NLLB-600M and mBART50-mmt. All models reach state-of-the-art performance on English-Chinese text translation.

The evaluation was done in two parts. We tested the performance of the fine-tuned models for translating general text with the Flores+ (NLLB Team et al., 2024; Costa-Jussà et al., 2022) and Tatoeba (Tiedemann, 2020) datasets. BLEU (Papineni et al., 2002), chrF2 (Popović, 2017) and CometKiwi (Rei et al., 2022) metrics were used to evaluate the translation. After that, we used the BEI dataset test split to see whether the fine-tuned models had learned about the negative semantic prosody of BEI passives and if they had a higher accuracy deciding when to use BEI passives in translation.

Finally, a probing task was conducted to assess whether the pretrained and fine-tuned models possess information that aids in correctly using the active or passive voice when translating English passives into Chinese. The task also aimed to identify which layers of the encoder or decoder store this information.

### 4.1 Model Fine-tuning

We fine-tuned OPUS-MT, NLLB-600M and mBART50-mmt with the BEI dataset for English-Chinese text translation. The BEI dataset was split into 75% training, 11.25% validation and 13.75% test. We conducted a hyperparameter search for the learning rate and used batches of 32. For OPUS-MT, we trained with a learning rate of $10^{-5}$ and acquired the model with the best validation BLEU after 6 epochs. For NLLB-600M, we trained with a learning rate of $5 \times 10^{-4}$ for 5 epochs, and kept the checkpoint with the best validation BLEU (at step 60). For mBART50-mmt, the appropriate learning rate is also $10^{-5}$. We trained for 5 epochs and at step 120 we obtained the model with the best BLEU.

### 4.2 Model Performance Analysis

#### 4.2.1 General Text Translation

In Table 1, we can see that the fine-tuned OPUS-MT achieved slightly higher BLEU and chrF2 scores on the Tatoeba dataset, fine-tuned NLLB-600M achieved a higher BLEU on Flores+ and a higher CometKiwi on Tatoeba, and fine-tuned mBART50-mmt achieved a higher CometKiwi on Tatoeba dataset. In general, after fine-tuning, the models maintained their original accuracy in the English-Chinese text translation task.

#### 4.2.2 BEI Test Set Translation

The BEI test set contains 65 positive evidence sentence pairs and 59 negative evidence sentence pairs. The performance of pretrained and fine-tuned models is shown in Table 2. Generally speaking, all three models tend to use BEI passives to translate BE passives, with the positive evidence test set yielding higher accuracy than the negative evidence test set. All three models achieved higher accuracy in using BEI passives when translating BE passives after fine-tuning. For the positive evidence test set, fine-tuned mBART50-mmt shows the highest accuracy, which means it used BEI passives to translate BE passives with unfavourable content in most cases. For the negative evidence dataset, fine-tuned NLLB-600M performs the best,

| Models | Flores+ | | | Tatoeba | | |
|---|---|---|---|---|---|---|
| | BLEU | chrF2 | CometKiwi | BLEU | chrF2 | CometKiwi |
| OPUS-MT | **32.1** | **21.6** | 83.9 | 32.7 | 21.9 | 79.5 |
| Fine-tuned OPUS-MT | 31.7 | 21.3 | **84.3** | **33.7** | **22.7** | **80.9** |
| NLLB-600M | 23.3 | **17.1** | **82.3** | **27.7** | **18.4** | 69.7 |
| Fine-tuned NLLB-600M | **24.4** | 16.4 | 80.4 | 25.3 | 18.1 | **74.7** |
| mBART50-mmt | **32.8** | **22.6** | **84.8** | **33.9** | **22.8** | 81.6 |
| Fine-tuned mBART50-mmt | 32.3 | 21.4 | 84.7 | 33.2 | 22.4 | **82.0** |

Table 1: BLEU, chrF2 and CometKiwi scores (in %) of pretrained and fine-tuned models on Flores+ and Tatoeba datasets (English to Chinese). Fine-tuning leads to a minor drop in general translation quality, the models retain accuracy close to the original ones.

avoiding BEI passives in the translation of BE passives with neutral or favourable content.

Since in Chinese it is always acceptable to use the active voice for all kinds of content in Chinese, it is more important to achieve high accuracy in the negative evidence test set. Examples (6) and (7) show how fine-tuned models avoid using BEI passives when translating a BE passive stating a neutral or favourable event:

(6) **Source text:** Oh yes, and I **have been told** they played all sorts of mad pranks.

**Target text:** 有的。人家和我说，他们做了好多发疯似的把戏。

'Yes. They **have told** me that they played many mad pranks.'

**OPUS-MT:** 哦，是的，我**被**告知他们玩各种疯狂的恶作剧。

'Oh yes. I **have been told** that they played all sorts of mad pranks.'

**Fine-tuned OPUS-MT:** 哦，是的，有人告诉我，他们玩各种疯狂的恶作剧。

'Oh yes. Someone **has told** me that they played all sorts of mad pranks.'

(7) **Source text:** You **were treated** as a son in my friend's house.

**Target text:** 你在我朋友家里是待你同儿子一样的。

"In my friend's house they **treated** you as a son."

**NLLB-600M:** 在我朋友的家里，你**被**当作儿子。

'In my friend's house, you **were treated** as a son.'

**Fine-tuned NLLB-600M:** 你在我朋友的家里就像一个儿子一样。

'In my friend's house you **were** like a son.'

### 4.2.3 Transference to Spanish-Chinese

Since NLLB-600M is a multilingual model, we hypothesize that the knowledge regarding the negative semantic prosody of BEI passives can be transferred and applied when translating passive sentences from other languages into Chinese. According to data from CORPES (*Corpus del Español del Siglo XXI*[6]), like in English, Spanish passive structures *SER*/*ESTAR* + past participle also have neutral semantic prosody (appearing in neutral contexts in around 78% of all cases). We had the BEI test set translated into Spanish by a native Spanish speaker and guaranteed that all the passive structures were preserved. Pretrained and fine-tuned NLLB-600M performance on the Flores+ and Tatoeba dataset (Spanish-Chinese), as well as BEI test set in Spanish, is shown in Table 3. Due to the unsolved issue of producing approximately 50% its output in Englsih when asked to translate Spanish to Chinese, mBART50-mmt performs poorly on this task and is not discussed here[7].

After fine-tuning with the English-Chinese BEI dataset, NLLB-600M achieved higher accuracy on both positive and negative evidence sentences in the BEI test set translated into Spanish. This indicates that knowledge of the negative semantic prosody of BEI passives learned from English-Chinese fine-tuning can be applied to other language pairs. As for the Spanish-Chinese translation performance, NLLB-600M maintained it on Flores+, but it worsened slightly on Tatoeba.

---
[6]Available at: https://www.rae.es/corpes/
[7]Discussion raised at: https://huggingface.co/facebook/mbart-large-50-many-to-many-mmt/discussions/13.

| BEI test set | OPUS-MT | Fine-tuned OPUS-MT | NLLB-600M | Fine-tuned NLLB-600M | mBART50-mmt | Fine-tuned mBART50-mmt |
|---|---|---|---|---|---|---|
| positive evidence | 75.4 | 81.5 | 69.2 | 78.5 | 81.5 | **84.6** |
| negative evidence | 10.2 | 25.4 | 35.6 | **59.3** | 35.6 | 50.8 |

Table 2: Accuracy (in %) of pretrained and fine-tuned models on BEI test set (English to Chinese).

| Model | Flores+ | | | Tatoeba | | | Spanish-Chinese BEI test set | |
|---|---|---|---|---|---|---|---|---|
| | BLEU | chrF2 | CometKiwi | BLEU | chrF2 | CometKiwi | pos-evi | neg-evi |
| NLLB-600M | 16.8 | **13.1** | **65.8** | **35.1** | **23.6** | **81.0** | 66.2 | 28.8 |
| Fine-tuned NLLB-600M | **17.9** | 13.0 | 65.5 | 29.5 | 21.9 | 78.3 | **80.0** | **59.3** |

Table 3: Performance (in %) of pretrained and fine-tuned NLLB-600M model on Flores+, Tatoeba and the Spanish-Chinese BEI test set (transfer learning).

### 4.3 Probing Task

Probing classifiers have been used to measure and interpret how certain linguistic information is encoded in deep neural networks. The idea is to train a classifier (a probe) to predict a linguistic property from a model representation (Belinkov, 2022). In our case, we use probing to determine whether MT models store information in their representations, and whether such information helps to decide on the use of the passive voice in the Chinese translation of a BE passive.

The probe input consists of the hidden state of each layer of the models, including encoder layers and decoder layers. Our probing classifier has a single linear layer initialized with the same set of weights for all models and layers. The probing task is a simple classification task. We built a probing dataset with the validation and test set of the BEI dataset, which includes 225 BE passives, and a number indicating whether the translation should be an active or passive sentence (0 for active and 1 for passive). We used 80% of the probing dataset for training the probe, and the remaining 20% for testing probing accuracy. We trained a probe and recorded its accuracy on the probing test set for each layer of all three pretrained models and our fine-tuned models. This revealed which layer encodes the most information that could help models decide whether active or passive voice is more appropriate for the translation. The probing results are shown in Figure 2.

NLLB-600M and mBART50-mmt models have the same architecture. Both consist of a 12-layer encoder and a 12-layer decoder. In the encoders, we can observe that deeper layers tend to achieve higher accuracy on the probing task. Fine-tuning improved the mean accuracy of the encoder probes on mBART50-mmt by approximately 3%, but for NLLB-600M, it did not make a great difference. As for the decoders, it seems that the linguistic information that helps choose the correct voice in Chinese translation is distributed equally in all layers. The probes for the decoders of both models reached a higher mean accuracy after fine-tuning (an approximate 4.5% improvement). OPUS-MT has a 6-layer encoder and a 6-layer decoder, with a smaller layer size compared to the other two models[8]. In the encoder, fine-tuning had almost no effect on probing accuracy. We observe that layer 3 achieved a higher accuracy compared to all other layers. As for the decoder, fine-tuning caused layers 3 and 4 to contain less linguist information related to our task, while giving more information to the last layer.

Overall, with input from the NLLB-600M and mBART50-mmt layers, the probe shows a better ability to distinguish those BE passives that should be translated into Chinese active sentences from those that can keep their passive voice, whereas layers of OPUS-MT contain less information. This is in line with the performance of the models on the BEI test set and may suggest that models with more layers have a higher ability to capture information related to semantic prosody.

---
[8]The hidden size of NLLB-600M and mBART50-mmt is 1024, while for OPUS-MT it is 512.

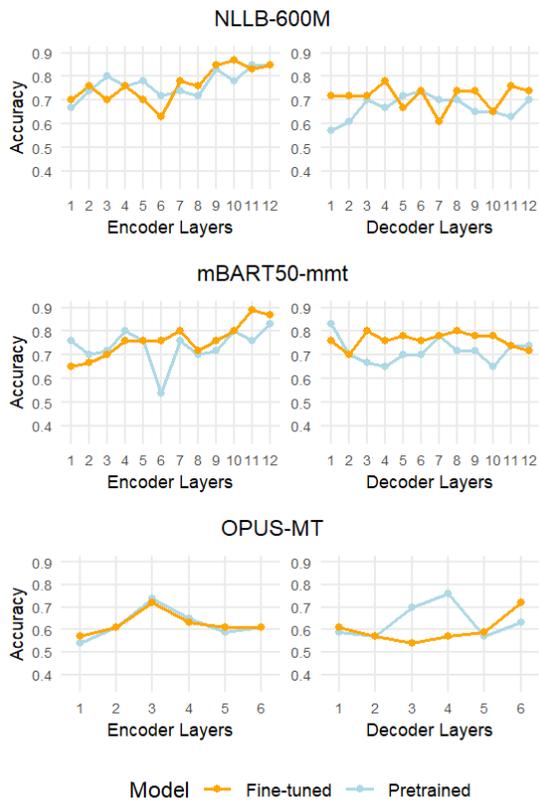

Figure 2: Layer-wise probing accuracy of pretrained and fine-tuned models. Encoder probes for NLLB-600M and mBART50-mmt show higher accuracy in deeper layers. Fine-tuning improves probe accuracy, particularly for the decoder layers. OPUS-MT encoders contain less task-relevant information overall, with accuracy peaking at layer 3.

## 5 Discussion

After fine-tuning the different models with the BEI dataset, they all showed improvements in reserving Chinese BEI passives for negative contexts. Here, we have a closer look at those cases with negative evidence that the model managed to translate correctly into the active voice, in order to find potential methods to further improve model performance.

The cases that models corrected after fine-tuning can be categorized into two kinds according to the translation strategy. For some BE passives, models switched the position of the subject and the object and removed the passive marker—such as in example (6)—in order to give a sentence in active voice. This is the kind of change we expected to see. The second translation strategy is to use a notional passive, which is also called a topic sentence, as in example (8):

(8) **Source text:** That they **have been preserved** so well.
**Target text:** 它们都保存得十分完整。
'They are in perfect preservation.'

**mBART50-mmt:** 它们**被**保存得如此好。
'They **have been preserved** so well.'

**Fine-tuned mBART50-mmt:** 它们保存得很好。
'They are in very good preservation.'

In a topic sentence, the subject argument is the topic and the patient of the verb (*THEY* in example (8)), while the remaining constituent is the comment. Topic sentences can express passive meaning because of the nature of their subjects, and sometimes it is grammatical to add a passive marker and turn it into a marked passive sentence, depending mainly on the verb. However, notional passives are only compatible with a very restricted number of verbs, and are not considered passive constructions in a strict sense (Tang, 2003).

In modern Mandarin Chinese literature, notional passives mainly occur in neutral contexts (over 80% of the time) and are twice as frequent as BEI passives (Guo and Chow, 2013), which means it is a plausible choice for translating BE passives into Chinese. Moreover, in example (8) we can observe that translating a BE passive into a notional passive does not require the model to change the order of subject and verb or to add back the omitted agent of the verb. The fine-tuned model only had to drop *BEI* 被 between the patient *THEY* 它们 and verb *PRESERVE* 保存 to obtain a topic sentence. In this case, *THEY* 它们 refers to non-human arguments, which means *THEY* cannot conduct the action of preserving something, but can only be the patient of the verb *PRESERVE*. Without the need to add the passive marker *BEI*, the sentence already transfers the passive notion that *THEY* are *PRESERVED*.

Using topic sentences to translate BE passives requires less movement and can preserve the passive meaning and neutral semantic prosody at the same time. However, while the BEI passive is more frequent in translated Chinese literature than in original Chinese literature, the opposite is the case with the notional passive (Guo and Chow, 2013). Both human translators and machine translation models should pay more attention to this structure to generate more accurate Chinese translations for passive sentences.

## 6 Conclusion and Future Work

In this work, we propose an approach for teaching sequence to sequence machine translation models about the semantic prosody of a specific structure, namely the Chinese BEI passive, and to improve model performance on translating English BE passives into Chinese. The primary focus of our approach is to fine-tune MT models with a dataset that explicitly demonstrates the negative semantic prosody of BEI passives through the contrast with using BEI passives to translate BE passives with negative context, while for BE passives with neutral and positive content, the translations are in the active voice. After fine-tuning OPUS-MT, NLLB-600M and mBART50-mmt, all models showed improvements in using BEI passives correctly in translation while maintaining their original performance on general text translation, showing that our approach is a valid one. Moreover, for multilingual models such as NLLB-600M, the knowledge of semantic prosody is transferable to translation tasks in other language pairs (e.g. from English-Chinese translation to Spanish-Chinese translation).

Through probing experiments, we found that, for NLLB-600M and mBART50-mmt, information that helps to decide whether the active or passive voice is more plausible in translation is concentrated in the deeper layers of the encoder and equally distributed in the decoder. Fine-tuning improved the probing accuracy of decoders significantly, but did not have a great influence on the encoders.

Our work focused on the BEI structure. A potential future research direction will be fine-tuning a model with combined data that can show the semantic prosody of multiple linguistic units and observe whether the model can learn different semantic prosodies simultaneously. Having a more diverse dataset can also prevent the problem of overfitting. However, there is no comprehensive list of linguistic units from different languages with semantic prosody that may cause inequivalence in translation. It would be of great value to compile such a list and create datasets that demonstrate their semantic prosody.

In English and Spanish, passive structures can be categorized into two different kinds, namely adjectival and verbal passives. In this work, we did not distinguish between them, as they present the same problem when translated into Chinese by MT models. However, we plan to investigate further to determine whether model performance varies on different kinds of passives.

Finally, another avenue for future work would be multilingual fine-tuning, which may yield better results on multilingual models. The difficulty here lies in the limited numbers of annotated multilingual parallel corpora, without which it would be difficult to collect data on a specific structure.

## Limitations

Although our dataset is enough to demonstrate the negative semantic prosody of Chinese BEI passives, there is another important fact that our dataset does not show. That is, the low frequency of passive sentences (including BEI passives and others) in general Chinese texts, which should also affect the usage of BEI passives in translation. Since all the models are pretrained models, they should have seen this fact in the training process already. Whether there is necessity of fine-tuning models with BEI dataset mixed in general text needs future work to justify.

## Acknowledgements

This work was supported by FairTransNLP-Language: Analysing toxicity and stereotypes in language for unbiased, fair and transparent systems (PID2021-124361OB-C33), funded by Ministerio de Ciencia, Innovación y Universidades, programa de I+D de Generación de Conocimiento (MICIU/AEI/10.13039/501100011033/FEDER,UE) and by Support grants to departments and university research units for the recruitment of pre-doctoral research staff in training (FI SDUR 2024).